\documentclass[11pt,a4paper]{article}
\usepackage[hyperref]{acl2020}
\usepackage{times}
\usepackage{latexsym}

\usepackage{microtype}
\usepackage{bm}
\usepackage{amsmath,amssymb,amsfonts}
\usepackage{enumerate}
\usepackage{algorithm}
\usepackage{algorithmic}
\usepackage{subfigure}
\usepackage{graphicx}
\usepackage{multirow}
\usepackage{booktabs}
\usepackage{CJKutf8}
\usepackage{arydshln}

\makeatletter
\newcommand\argmax{\mathop{\operator@font argmax}}
\newcommand\argmin{\mathop{\operator@font argmin}}
\makeatother

\aclfinalcopy % Uncomment this line for the final submission
\title{Cross-Lingual Semantic Role Labeling with High-Quality Translated Training Corpus}

\author{
Hao Fei$^1$ \and Meishan Zhang$^{2}$\thanks{~~Corresponding author.} \and Donghong Ji$^1$ \\
1. Department of Key Laboratory of Aerospace Information Security and Trusted Computing,\\
Ministry of Education, School of Cyber Science and Engineering, Wuhan University, China \\ 
2. School of New Media and Communication, Tianjin University, China \\
\texttt{\{hao.fei,dhji\}@whu.edu.cn} \and \texttt{mason.zms@gmail.com} \\
}

\date{}

\begin{document}
\maketitle
\begin{abstract}
Many efforts of research are devoted to semantic role labeling (SRL) which is crucial for natural language understanding.
Supervised approaches have achieved impressing performances when large-scale corpora are available for resource-rich languages such as English.
While for the low-resource languages with no annotated SRL dataset,
it is still challenging to obtain competitive performances.
Cross-lingual SRL is one promising way to address the problem,
which has achieved great advances with the help of model transferring and annotation projection.
In this paper, we propose a novel alternative based on corpus translation,
constructing high-quality training datasets for the target languages from the source gold-standard SRL annotations.
Experimental results on Universal Proposition Bank show that
the translation-based method is highly effective,
and the automatic pseudo datasets can improve the target-language SRL performances significantly.
\end{abstract}

\section{Introduction}
Semantic role labeling (SRL), which aims to capture the high-level meaning of a sentence,
such as \emph{who did what to whom}, is an underlying task for facilitating a broad range of natural language processing (NLP) tasks \cite{shen-lapata-2007-using,liu-gildea-2010-semantic,genest-lapalme-2011-framework,gao-vogel-2011-corpus,wang-etal-2015-machine,KhanSK15}.
Currently, the majority of research work on SRL is dedicated to the English language, due to the availability of large quantity of labeled data.
With this regard, cross-lingual SRL, especially the one transferring the advantage of the source language with affluent amount of resources (e.g., English) to the target language where the labeled data is scarce or even not available, is of great importance \cite{KozhevnikovT13,he-etal-2019-syntax,aminian-etal-2019-cross}.

\begin{figure}[!t]
\centering
\includegraphics[width=.84\columnwidth]{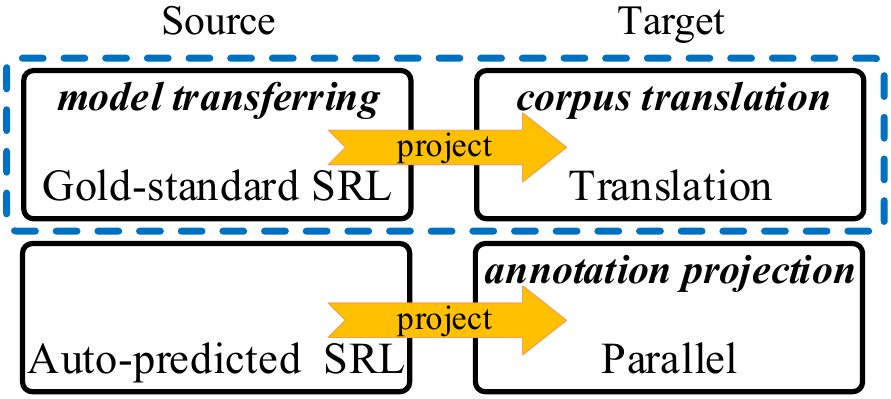}
\caption{
Illustration of cross-lingual SRL methods.
Our method is in the dotted blue box.
}
\label{methods}
\end{figure}

Previous work on cross-lingual SRL can generally be divided into two categories: model transferring and annotation projection.
The former builds cross-lingual models on language-independent features such as cross-lingual word representations and universal POS tags which can be transferred into target languages directly   \cite{mcdonald-etal-2013-universal,swayamdipta-etal-2016-greedy,daza-frank-2019-translate}.
The latter bases on a large-scale parallel corpus between the source and target languages where the source-side sentences are annotated with SRL tags automatically by a source SRL labeler,
and then the source annotations are projected onto the target-side sentences in accordance of word alignments \cite{yarowsky-etal-2001-inducing,HwaRWCK05,van-der-plas-etal-2011-scaling,KozhevnikovT13,Pado2014Cross,akbik2015generating}.
In addition, the annotation projection can be combined with model transferring naturally.

Particularly, the projected SRL tags in annotation projection could contain much noise because of the source-side automatic annotations.
A straightforward solution is the translation-based approach,
which has been demonstrated effective for cross-lingual dependency parsing \cite{TackstromMU12,RasooliC15,GuoCYWL16,zhang-etal-2019-cross}.
The key idea is to translate the gold-standard source training data into target language side by translation directly,
avoiding the problem of the low-quality source annotations.
Fortunately, due to recent great advances of neural machine translation (NMT) \cite{BahdanauCB14,WuSCLNMKCGMKSJL16},
this approach could have great potentials for cross-lingual transferring.

To this end, in this paper, we study the translation-based method for cross-lingual SRL.
Figure \ref{methods} illustrates the differences between previous approaches.
Sentences of the source language training corpus are translated into the target language,
and then the source SRL annotations are projected into the target side,
resulting in a set of high-quality target language SRL corpus,
which is used to train the target SRL model.
Further, we merge the gold-standard source corpus and the translated target together,
which can be regarded as a combination of the translation-based method and the model transferring.
Our baseline is a simple BiLSTM CRF model by using multilingual contextualized word representations \cite{PetersNIGCLZ18,devlin-etal-2019-bert}.
For a better exploration of the blended corpus,
we adopt a parameter generation network (PGN) to enhance the BiLSTM module,
which can capture the language differences effectively \cite{platanios-etal-2018-contextual,jia-etal-2019-cross}.

We conduct experiments based on Universal Proposition Bank corpus (v1.0) \cite{akbik2015generating,akbik2016polyglot} over seven languages.
First, we verify the effectiveness of our method on the single-source  SRL transferring,
where the English language is adopted as the source language and the remaining are used as the target languages.
Results show that the translation-based method is highly effective for cross-lingual SRL,
and the performances are further improved when PGN-BiLSTM is used.
Further, we conduct experiments on the multi-source SRL transferring,
where for each target language all the remaining six languages are used as the source languages.
The same tendencies as the single-source setting can be observed.
We conduct detailed analysis work for both settings to understand our proposed method comprehensively.

In summary, we make the following two main contributions in this work:
\begin{itemize}
    \item We present the first work of the translation-based approach for unsupervised cross-lingual SRL. 
    We build a high-quality of pseudo training corpus for a target language, and then verify the effectiveness of the corpus under a range of settings.
    \item We take advantage of the multilingual contextualized word representations, and strengthen the multilingual model training with PGN-BiLSTM model.
\end{itemize}
All codes and datasets are released publicly available for the research purpose\footnote{\url{https://github.com/scofield7419/XSRL-ACL} under Apache License 2.0.}.

\section{Related Work}

There exists extensive work for cross-lingual transfer learning \cite{van-der-plas-etal-2011-scaling,KozhevnikovT13,Pado2014Cross,RasooliC15,TiedemannA16,0002ZWCMG18,ChenSACW18,chen-etal-2019-multi-source,aminian-etal-2019-cross}.
Model transferring and annotation projection are two mainstream categories for the goal.
The first category aims to build a model based on the source language corpus,
and then adapt it to the target languages \cite{yarowsky-etal-2001-inducing,HwaRWCK05,tiedemann-2015-improving}.
The second category attempts to produce a set of automatic training instances for the target language by a source language model and a number of parallel sentences,
and then train a target model on the dataset \cite{bjorkelund-etal-2009-multilingual,mcdonald-etal-2013-universal,lei-etal-2015-high,swayamdipta-etal-2016-greedy,mulcaire-etal-2018-polyglot,daza-frank-2019-translate}.

For cross-lingual SRL,
annotation projection has received the most attention \cite{Pado2014Cross}.
A range of strategies have been proposed to enhance the SRL performance of the target language,
including improving the projection quality \cite{tiedemann-2015-improving},
joint learning of syntax and semantics \cite{KozhevnikovT13},
iterative bootstrapping to reduce the influence of noise target corpus \cite{akbik2015generating},
and joint translation and SRL \cite{aminian-etal-2019-cross}.

Our work is mainly inspired by the recent work of treebank translation of cross-lingual dependency parsing \cite{tiedemann-etal-2014-treebank,tiedemann-2015-improving,RasooliC15,GuoCYWL16,TiedemannA16,conneau-etal-2018-xnli,zhang-etal-2019-cross},
which is referred to as the translation-based approaches.
These approaches directly project the gold-standard annotation into the target side,
alleviating the problem of erroneous source annotations in standard annotation projection.
In addition, we combine the approach with model transferring,
which has been concerned little for cross-lingual SRL.
The model transferring benefits much from the recent advance of cross-lingual contextualized word representations \cite{he-etal-2019-syntax}.

The development of universal annotation schemes for a variety of NLP tasks can greatly facilitate cross-lingual SRL,
including POS tagging \cite{PETROV12-274}, dependency parsing \cite{mcdonald-etal-2013-universal,przepiorkowski-patejuk-2018-arguments}, morphology \cite{sylak-glassman-etal-2015-language} and SRL \cite{aminian-etal-2019-cross}.
Our work makes use of the publicly available Universal Proposition Bank (UPB) \cite{akbik2015generating,akbik2016polyglot},
which annotates the predicate and semantic roles following the frame and role schemes of the English Proposition Bank 3.0 \cite{KingsburyP02,Palmer2005}.

Supervised SRL models are also closely related to our work \cite{he-etal-2017-deep,he-etal-2018-jointly,XiaL0ZFWS19}.
A great deal of work attempts for an end-to-end solution with sophistical neural networks, detecting the predicates as well as the corresponding argument roles in one shot \cite{he-etal-2017-deep,TanWXCS18,LiHZZZZZ19}.
Also there exist a number of studies which aims for adapting various powerful features for the task \cite{strubell-etal-2018-linguistically,li-etal-2018-unified}.
In this work, we exploit a multilingual PGN-BiLSTM model \cite{jia-etal-2019-cross} with contextualized word representations \cite{he-etal-2019-syntax},
which can obtain state-of-the-art performance for cross-lingual SRL.

\section{SRL Translation}

We induce automatic target data from the gold-standard source data by full translation
and then project the SRL predicates and arguments into their corresponding words by aligning,
producing the final translated SRL corpus for the target language automatically.
The method has been demonstrated effective for cross-lingual dependency parsing \cite{tiedemann-etal-2014-treebank,tiedemann-2015-improving,TiedemannA16,zhang-etal-2019-cross}.
Compared with annotation projection,
we can ensure the annotation quality at the source side,
thus higher quality target corpus is also expected.
In addition, dependency-based SRL could benefit more by this method,
as only predicate words and their arguments are required to be projected into the target side,
while dependency parsing should concern all sentential words.
The overall process is accomplished by two steps: translating and projecting.

\begin{figure}[!t]
\centering
\subfigure[One-to-one projection.]{\label{Translationa} \includegraphics[scale=0.37]{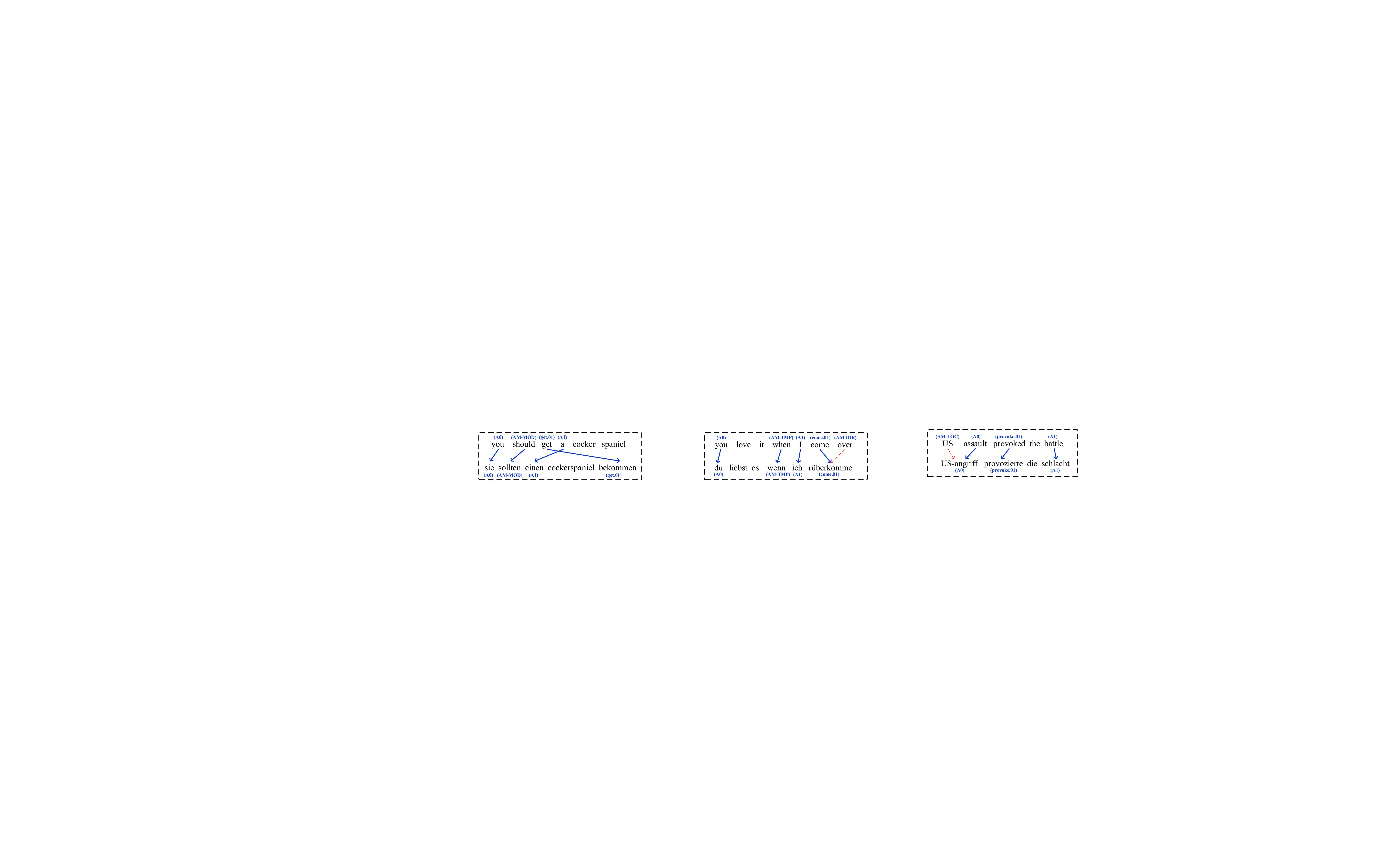}}
\subfigure[Predicate-argument collision. Only keep predicate.]{\label{Translationb} \includegraphics[scale=0.37]{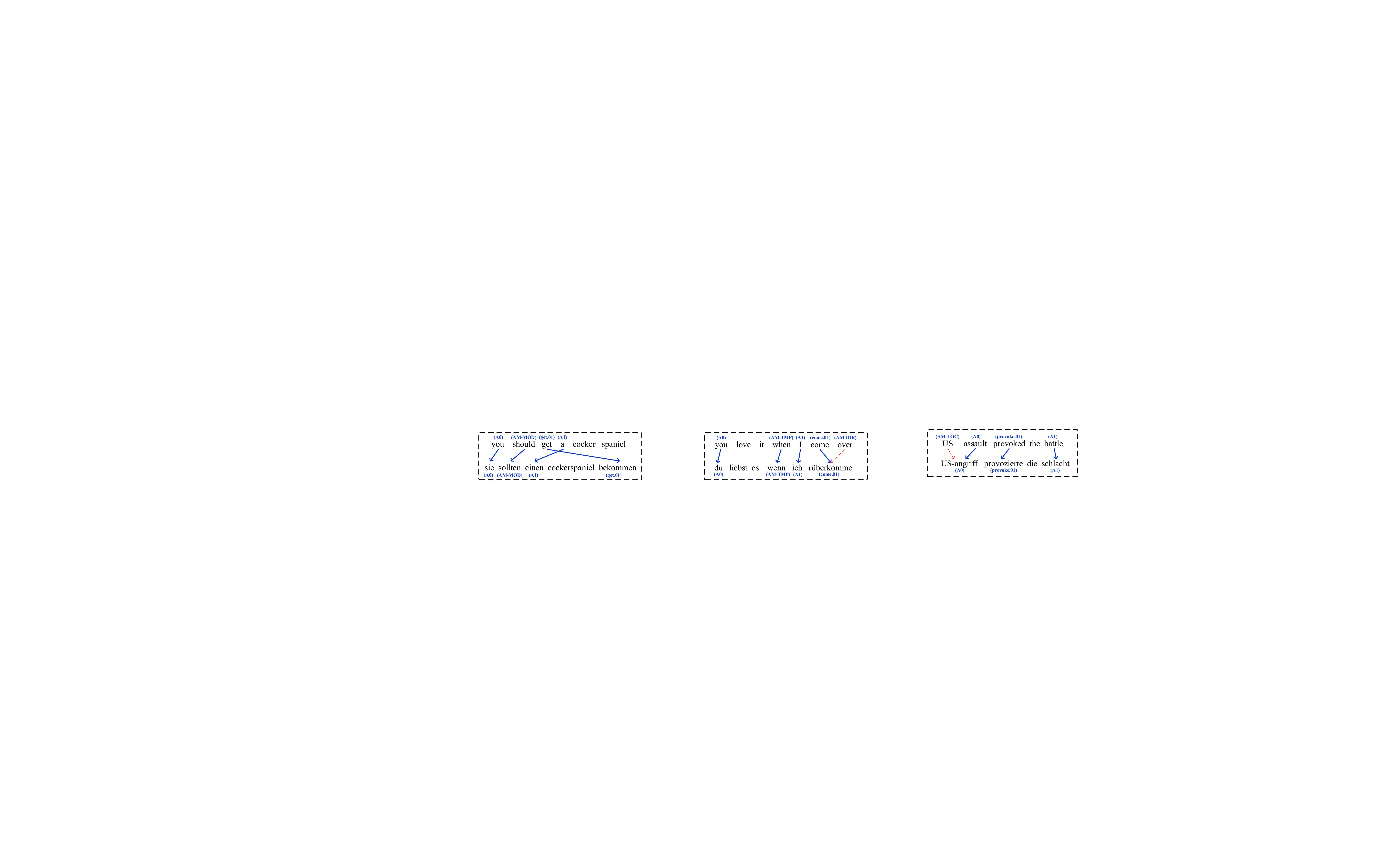}}
\subfigure[Argument-argument collision. Only keep the one with higher confidence.]{\label{Translationc} \includegraphics[scale=0.37]{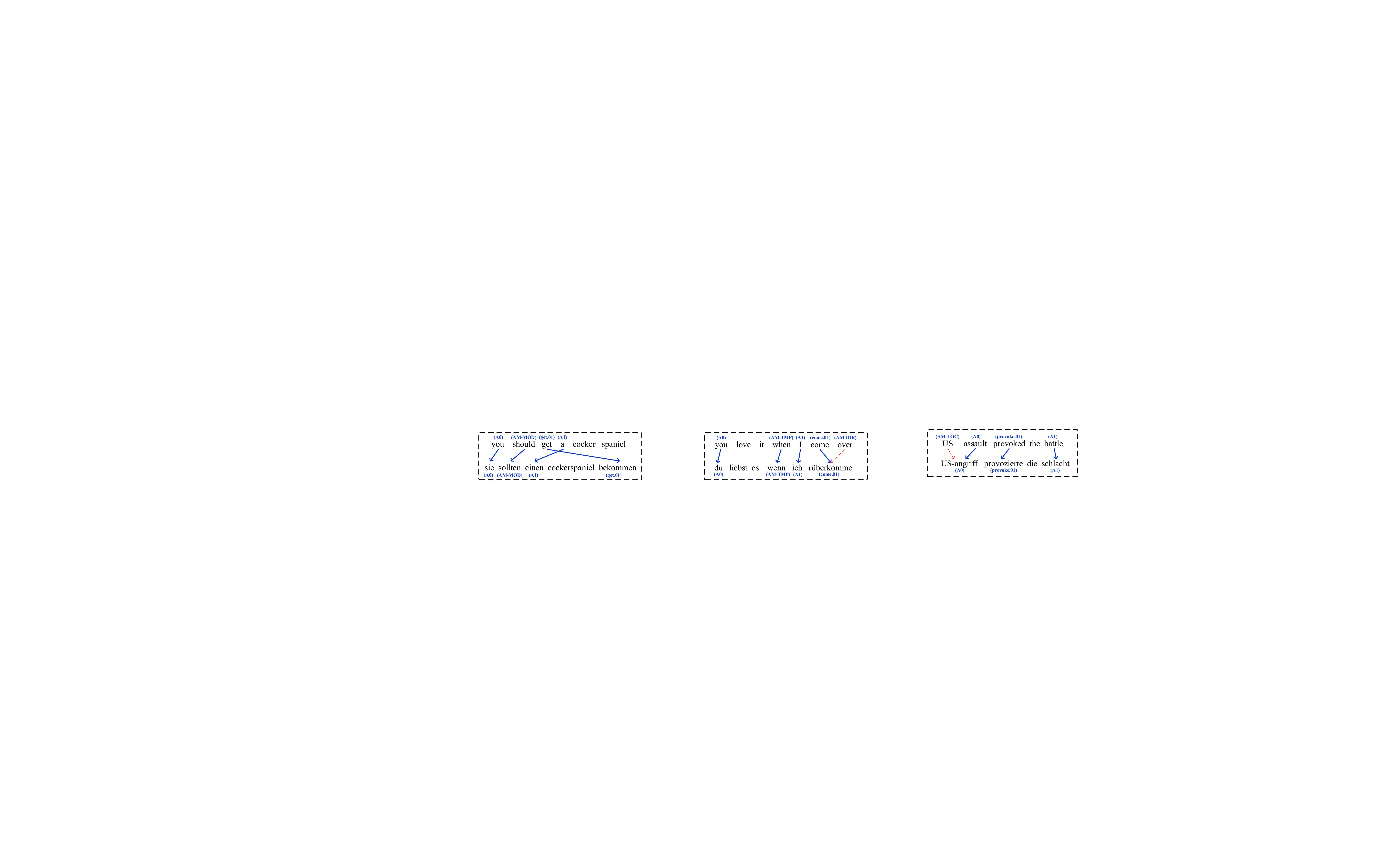}}
\caption{
Examples of SRL projection from English (upper) to German (lower),
where the thick blue solid arrows indicate successful projections, and the thin red dotted arrows indicate invalid projections.
}
\label{projection}
\end{figure}

\paragraph{Translating.}
First, we use a state-of-the-art translation system to produce the target translations for the sentences of the source SRL data.
Give a source sentence $e_1\cdots e_{n}$, we translate it into $f_1\cdots f_m$ of the target language.
It is worth noting that the recent impressive advances in NMT \cite{BahdanauCB14,WuSCLNMKCGMKSJL16}  facilitate our work greatly, 
which enables our method to have high-quality translations.

\paragraph{Projecting.}
Then we incrementally project the corresponding predicates or arguments of a source sentence $e_1\cdots e_{n}$  to its target $f_1\cdots f_m$.
We adopt two kinds of information to assist the projection:
(1) the alignment probabilities $a(f_j|e_i)$ from the source word $e_i$ into $f_j$ , which can be calculated by a word-alignment tool,
and (2) the POS tag distributions $p(t_{*} | f_j )$ of the target sentential words, which can be derived from a supervised target language POS tagger,
where $i \in [1, n]$, $j \in [1, m]$, and $t_*$ denotes an arbitrary POS tag.

We focus on SRL-related words of the source sentence only,
and perform the process gradually at the predicate level.
For each predicate in a sentence, we collect the predicate word as well as its role words,
and then project their role labels into the target sentence.
Formally, for each of these words (i.e., $e_i$), we have the SRL role tag $r_{e_i}$ as well as its POS tag $t_{e_i}$,
both of which have been already annotated in the UPB.
First, we find its target word $f_j$ with the highest alignment probability,
regarding the word $f_j$ as the corresponding projection carrying the semantic role $r_{e_i}$.
Then we calculate the confidence score of this projection by the following formula:
\begin{equation}
\mathrm{score}(e_i\rightarrow f_j, r_{e_i}) = a(f_j|e_i)p(t_{e_i}|f_j),
\end{equation}
which is a joint probability of word alignment corresponding and POS tag consistency.

The one-one target-source alignment \ref{Translationa} is the ideal condition of the projection.
However, there could be many-to-one cases for the given words,
leading to semantic role conflicts at the target language words.
For these cases, we take precedence for the predicate projections,
and otherwise keep only the highest confidence projections.
Figure \ref{Translationb} shows a predicate-argument conflict example, where the predicate projection is reserved, 
and Figure \ref{Translationc} shows an argument-argument conflict example where the projection with the higher confidence score is reserved.

Finally, we set a threshold value $\alpha$ to remove low confidence projections.
If the confidence score of a predicate projection is below $\alpha$,
all the roles of this predicate are removed as well.
For the argument projections whose confidence is below $\alpha$, we remove the single arguments directly,
with no influence on the other projections.

\section{The SRL Model}
\begin{figure}[!t]
\centering
\includegraphics[width=.99\columnwidth]{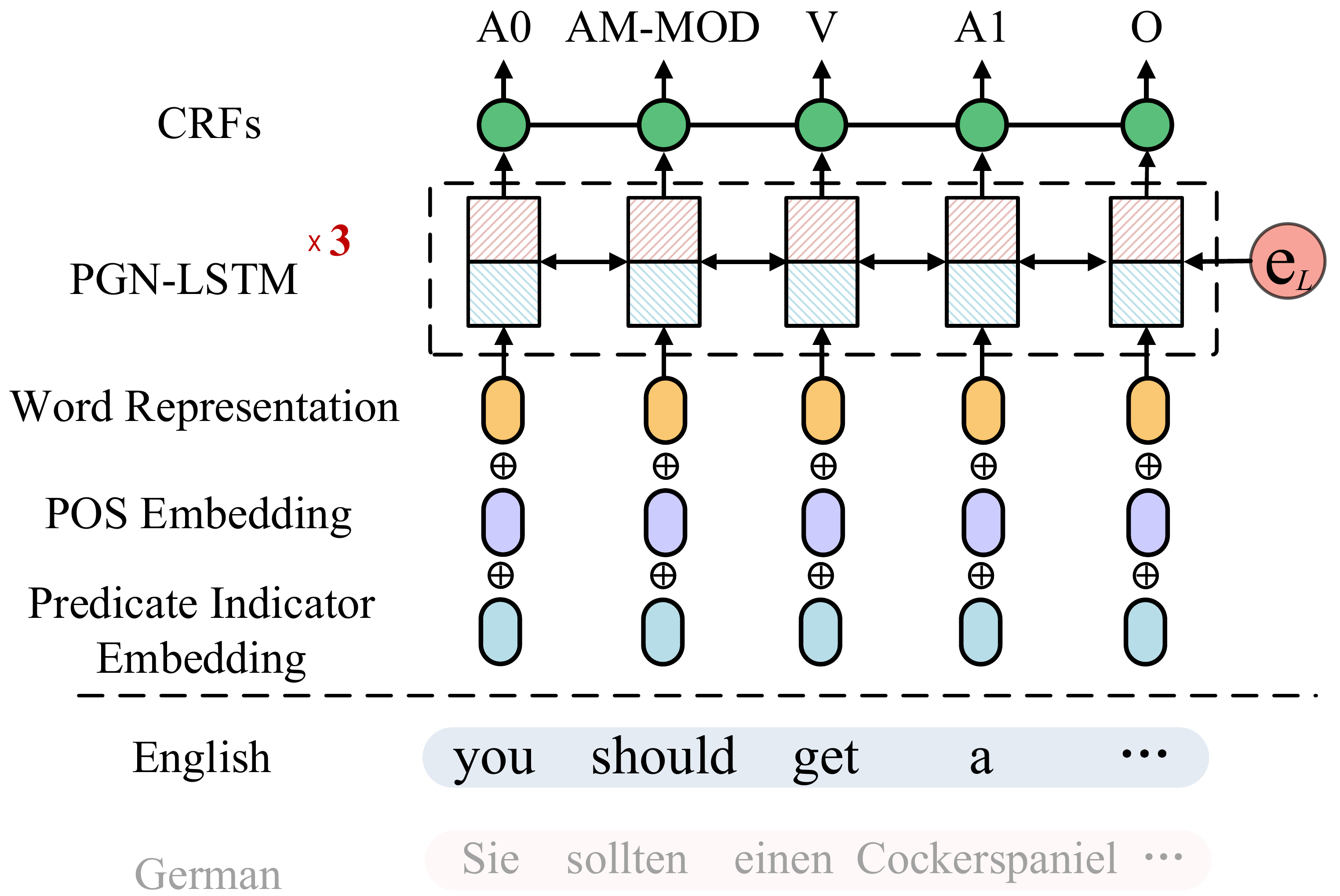}
\caption{
The overall architecture of the SRL model.
}
\label{encoder}
\end{figure}

In this work, we focus on dependency-based SRL, recognizing semantic roles for a given predicate \cite{he-etal-2017-deep}.
The task can be treated as a standard sequence labeling problem,
and a simple multi-layer BiLSTM-CRF model is exploited here,
which has archived state-of-the-art performance with contextualized word representations \cite{he-etal-2018-syntax,XiaL0ZFWS19,he-etal-2019-syntax}.
In particular, we adapt the model to better support multilingual inputs by using a PGN module on the BiLSTM \cite{hochreiter1997long}.
Figure \ref{encoder} shows the overall architecture.

\subsection{Word Representation}
\label{methodWordRep}
Given an input sentence $s = w_{1}\cdots w_{n}$ of a specific language $\mathcal{L}$ and $w_p$ ($p$ denotes the position) is the predicate word,
we use three sources of features to represent each word: (1) the word form, (2) the POS tag and (3) the predicate indicator:
\begin{equation}
    \bm{x}_{i} = \bm{v}_{w_i} \oplus \bm{v}_{t_i} \oplus \bm{v}_{(i==p)},
\end{equation}
where $t_1 \cdots t_n$ is the universal POS tag sequence for the input sentence.
For the POS tags and the predicate indicators, we use the embedding method to obtain their vectorial representations.

We compare three kinds of word form representations for cross-lingual SRL:
(1) multilingual word embeddings,
(2) multilingual ELMo representation \cite{PetersNIGCLZ18},
and (3) multilingual BERT representation \cite{devlin-etal-2019-bert}.
Note that we use the averaged vector of the inner-word piece representations from BERT outputs as the full word representation.

\subsection{Encoding Layer}
We employ the PGN-BiLSTM \cite{platanios-etal-2018-contextual,jia-etal-2019-cross} to encode the input sequence $\bm{x}_{1} \cdots \bm{x}_{n}$,
which is first introduced for cross-domain transfer learning to capture domain difference.
Here we use it for the multilingual setting aiming to model the language characteristics.

\begin{figure}[!t]
\centering
\includegraphics[width=0.70\columnwidth]{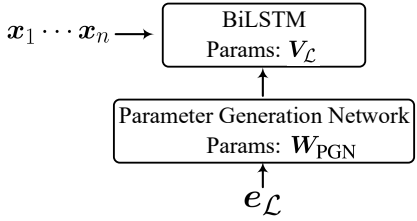}
\caption{
The mechanism of the PGN-BiLSTM.
}
\label{PGN}
\end{figure}

Compared with the vanilla BiLSTM module, PGN-BiLSTM dynamically selects the language-aware parameters for BiLSTM.
Let $\bm{V}$ be the flattened vector of all the parameters of a BiLSTM cell,
the language-aware $\bm{V}_{\mathcal{L}}$ is produced by:
\begin{equation}
    \bm{V}_{\mathcal{L}} = \bm{W}_{\text{PGN}} \times \bm{e}_{\mathcal{L}},
\end{equation}
where $\bm{W}_{\text{PGN}}$ denotes the parameters of vanilla BiLSTM part in the PGN-BiLSTM, including the weights of the input, forget, output gates and the cell modules, and $\bm{e}_{\mathcal{L}}$ is the embedding representation of language $\mathcal{L}$.
The mechanism of parameter generation of PGN-BiLSTM is illustrated in Figure \ref{PGN}.
Following, we derive module parameters from $\bm{V}_{\mathcal{L}}$ to compute the BiLSTM outputs.
The overall process can be formalized as:
\begin{equation}
\begin{split}
\bm{h}_{1} \cdots \bm{h}_{n} & = \text{PGN-BiLSTM}(\bm{x}_{1} \cdots \bm{x}_{n}, \bm{e}_{\mathcal{L}}) \\
    &= \text{BiLSTM}_{\bm{V}_{\mathcal{L}}} (\bm{x}_{1} \cdots \bm{x}_{n})
\end{split}
\end{equation}
which differs from the vanilla BiLSTM in that $\bm{e}_{\mathcal{L}}$ is one extra input to obtain BiLSTM parameters.
Specifically, we adopt a 3-layer bidirectional PGN-LSTM as the encoder.

\subsection{Output Layer}

Given the encoder output $\bm{h}_{1} \cdots \bm{h}_{n}$ for sentence $s = w_1 \cdots w_n$, we use CRFs \cite{LaffertyMP01} to
compute the probability of each candidate output $y = y_1\cdots y_n$:
\begin{equation}
\begin{split}
\bm{o}_i  &= \bm{W}\bm{h}_{n}, i \in [1, n] \\
p(y |s) &= \frac{\text{exp}\{ \sum_{i}(\bm{o}_{i} + \bm{T}_{y_{i-1}, y_{i}} )\}} {Z}
\end{split}
\end{equation}
where $\bm{W}$ and $\bm{T}$ are the parameters of CRFs, and $Z$ is a normalization factor for probability calculation.
The Viterbi algorithm is used to search for the highest-probability output SRL tag sequence.

\section{Experiments}

\subsection{Universal Proposition Bank}

Our experiments are based on the Universal Proposition Bank (UPB, v1.0) \footnote{\url{https://github.com/System-T/UniversalPropositions}}, which is built upon Universal Dependency Treebank (UDT, v1.4)\footnote{\url{https://lindat.mff.cuni.cz/repository/xmlui/handle/11234/1-1827}} and Proposition Bank (PB, v3.0)\footnote{\url{http://propbank.github.io/}}.
In UPB, consistent dependency-based universal SRL annotations are constructed across all languages.
In particular, we assemble the English SRL dataset based on the English EWT subset from the UDT v1.4 and the English corpus in PB v3.0.
Finally, we choose a total of seven languages as our datasets,
including English (EN) and German (DE) of the IE.German family, French (FR), Italian (IT), Spanish (ES)\footnote{We merge the Spanish and Spanish-AnCora as one.} and  Portuguese (PT) of the IE.Romance family, and Finnish (FI) of the Uralic family.
Table \ref{Dataset of UPB} shows the data statistics in detail.

\subsection{SRL Translation}

We focus on unsupervised cross-lingual SRL, assuming that no gold-standard target-language SRL corpus is available.
Our goal is to construct pseudo training datasets by corpus translation from the gold-standard source-language SRL datasets.
The Google Translation System \footnote{\url{https://translate.google.com/}, Oct. 1 2019} is adopted for sentence translation,
and the fastAlign toolkit \cite{dyer2013} is used to obtain word alignments.
In order to obtain accurate word alignment, we collect a set of parallel corpora to augment the training dataset of fastAlign.\footnote{\url{http://opus.nlpl.eu/}, Europarl v8.}
The universal POS tags of the translated sentences are produced by supervised monolingual POS taggers,
which are trained on the corresponding UDT v1.4 datasets, respectively.\footnote{A simple BiLSTM-CRF POS tagging model with monolingual ELMo representations is used, which can achieve accuracies of 96.54\%(EN), 97.15\%(DE), 94.42\%(FR), 97.21\%(IT), 94.12\%(ES), 95.86\%(PT) and 92.16\%(FI), respectively.  }

\begin{table}[!t]
\begin{center}
\resizebox{1.0\columnwidth}{!}{
  \begin{tabular}{ccrrrrr}
\hline
 \texttt{Fam.}  & \texttt{Lang.}  & \texttt{Train}  & \texttt{Dev} &  \texttt{Test}  & \texttt{Pred.}& \texttt{Arg.} \\
\cmidrule(r){1-1}\cmidrule(r){2-2}\cmidrule(r){3-5}\cmidrule(r){6-7}
\multirow{2}{*}{IE.Ge} & EN &10,907&	1,631&	1,633&	41,359&	100,170\\
 & DE & 14,118&	799&	977&	23,256&	58,319\\
\cmidrule(r){1-1}\cmidrule(r){2-2}\cmidrule(r){3-5}\cmidrule(r){6-7}
\multirow{4}{*}{IE.Ro} & FR &14,554&	1,596&	298&	26,934&	44,007\\
 & IT &12,837&	489&	489&	26,576&	56,893\\
 & ES &28,492&	3,206&	1,995&	81,318&	177,871\\
 & PT &7,494&	938&	936&	19,782&	41,449\\
\cmidrule(r){1-1}\cmidrule(r){2-2}\cmidrule(r){3-5}\cmidrule(r){6-7}
Ura & FI &12,217&	716&	648&	27,324&	60,502\\
\hline
\end{tabular}
}
\end{center}
  \caption{Statistics of the UPB, 
  where \texttt{Fam.} indicates the language famaily,
  IE.Ge refers to the Indo-European Germanic, IE.Ro refers to the Indo-European Romance,and Ura represents Uralic.
  }
  \label{Dataset of UPB}
\end{table}

\subsection{Settings}

\paragraph{Multi-lingual word representations.}
As mentioned in Section \ref{methodWordRep}, we investigate three kinds of multilingual word representations:
(1) Word Embedding (Emb): \texttt{MUSE} is exploited to align all monolingual fastText word embeddings into a universal space \cite{museLampleCRDJ18}.\footnote{\url{https://github.com/facebookresearch/MUSE}}
(2) ELMo: A blended dataset\footnote{CoNLL2017 corpus: \url{https://lindat.mff.cuni.cz/repository/xmlui/handle/11234/1-1989}} of the seven languages is used to train multilingual ELMo \cite{Crosslingual-elmo}.
(3) BERT: the official released multilingual BERT (base, cased version) is used directly \cite{devlin-etal-2019-bert}.\footnote{\url{https://github.com/google-research/bert}}

\begin{figure}[!t]
\centering
\includegraphics[width=1.0\columnwidth]{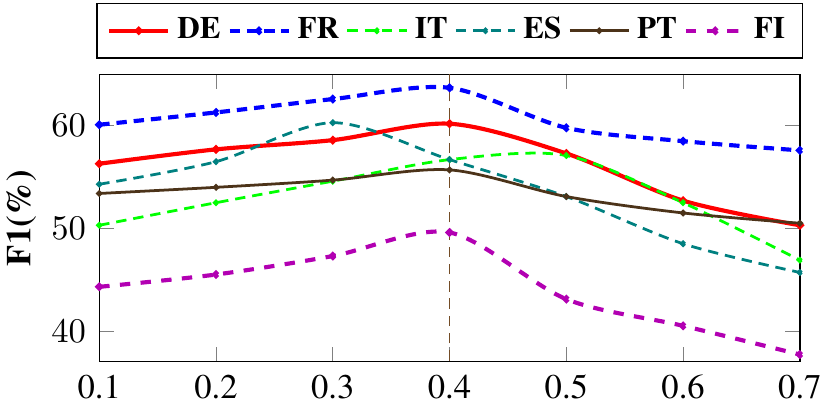}
\caption{
Performances with the translated target data under varying projection threshold $\alpha$.
}
\label{alpha}
\end{figure}

\paragraph{Hyperparameters.}
For SRL translation, there is only one hyperparameter, the projection confidence threshold $\alpha$, for filtering low-quality translated SRL sentences.
Figure \ref{alpha} shows the performances in the preliminary experiments for each languages under different $\alpha$.
Accordingly, we set $\alpha$ universally for all languages to $0.4$.
For the neural SRL models,
the dimension sizes of multilingual word embeddings, ELMo and BERT are 300, 1024 and 768, respectively.
The POS tag, predicate-indicator and language ID embedding sizes are 100, 100 and 32, respectively.
The hidden size of LSTM is set to 650.
We exploit online training with a batch size of 50,
and the model parameters are optimized by using the Adam algorithm with an initial rate of 0.0005.
The training is performed over the whole training dataset without early-stopping
for 80 iterations on bilingual transfer,
and 300 iterations on multilingual transfer.

\paragraph{Baselines.}
In order to test the effectiveness of our \emph{PGN} model,
we compare it with several baselines as well.
First, we denote our model by using the vanilla BiLSTM instead as \emph{BASIC},
and in particular, this model is exploited for all monolingual training all through this work.
Further, we adopt two much stronger baselines,
the \emph{MoE} model proposed by \newcite{guo-etal-2018-multi}
and the \emph{MAN-MoE} model proposed by \newcite{chen-etal-2019-multi-source}, respectively.
Both the two models are designed to train a model effectively based on corpora from multiple languages,
similar to our \emph{PGN} model.

\paragraph{Evaluation.}
We use the F1 score as the major metric to measure the model performance for each target language.
Each model is trained five times and the averaged value is reported.
We conduct significance tests by using the Dan Bikel’s randomized parsing evaluation comparator\footnote{\url{http://www.cis.upenn.edu/˜dbikel/software.html\#comparator}}.

\begin{table}[!t]
\begin{center}
\resizebox{1.0\columnwidth}{!}{
  \begin{tabular}{c|cccccc|c}
\hline
 Model & ~DE~& ~FR~& ~IT~& ~ES~& ~PT~& ~FI~ &\texttt{Avg}\\
\hline \hline
\multicolumn{8}{c}{\textbf{SRC}}\\ \hline
\rotatebox{0}{Emb} & 42.7 & 51.0 & 42.6 & 40.1 & 43.9 & 30.0 & 41.7 \\
\rotatebox{0}{BERT} &	43.2 & 53.1 & 44.4 & 41.2 & 44.2 & 31.6  & 43.0 \\
\rotatebox{0}{ELMo}&	46.8 & 54.6 & 43.0 & 42.1 & 46.1 & 33.9  & 44.4 \\
\hline\hline
\multicolumn{8}{c}{\textbf{TGT} }\\ \hline
\rotatebox{0}{Emb} &   49.4 & 51.3 & 45.5 & 48.4 & 46.9 & 38.7  & 46.7 \\
\rotatebox{0}{BERT} &  53.0 & 54.3 & 49.1 & 51.3 & 48.8 & 41.1  & 49.6 \\
\rotatebox{0}{ELMo} & 54.6 & 55.3 & 49.7 & 53.6 & 49.8 & 43.9  & 51.1 \\
\hline\hline
\multicolumn{8}{c}{\textbf{SRC \& TGT} ~~~(ELMo) } \\ \hline
\emph{BASIC} & 59.2 & 61.7 & 55.1 & 58.3 & 53.7 & 47.6  & 55.8 \\
\emph{PGN} & \textbf{65.0} & 64.8 & \textbf{58.7} & 62.5 & \textbf{56.0} & \textbf{54.5}  & \textbf{60.3} \\
\hdashline
\emph{MoE} & 63.2 & 63.3 & 56.7 & 60.3 & 55.0 & 50.6  & 58.2 \\
\emph{MAN-MoE} & 64.3 & \textbf{65.3} & 57.1 & \textbf{62.8} & 55.2 & 52.3  & 59.4 \\
\hline
\end{tabular}
}
\end{center}
  \caption{Results of cross-lingual transfer from English. }
  \label{Xling:En}
\end{table}

\subsection{Cross-Lingual Transfer from English}

We first conduct experiments on cross-lingual transfer from the English source to the rest of the other six target languages, respectively,
which has been a typical setting for cross-lingual investigations \cite{wang-etal-2019-cross}.
The results are summarized in Table \ref{Xling:En}.
We list the F-scores by using only the source corpus ({\textbf SRC}), only the translated target corpus ({\textbf TGT})
and the mixture corpus of source and target ({\textbf SRC \& TGT}),
comparing the performances of different multilingual word representations
as well as different multilingual SRL models.

\paragraph{Multilingual word representations.}
First, we evaluate the effectiveness of the three different multilingual word representations exploited.
We compare their performances under two settings, by using {\textbf SRC} and {\textbf TGT} corpus, respectively.
According to the results, we find that the multilingual contextualized word representations (i.e. BERT and ELMo) are better in both two settings,
which is consistent with previous studies \cite{Crosslingual-elmo,schuster-etal-2019-cross}.
Interestingly, the multilingual BERT performs worse than the ELMo,
which can be explained by that the ELMo representation is pre-trained based on the corpus which involves in the focused seven languages.
This indicates that the official released multilingual BERT can be further improved, 
since monolingual BERT has been demonstrated to produce better performances than ELMo \cite{Probing19}.

\paragraph{Translated target.}
Next, We consider taking the translated target as only the training data to examine the effectiveness of the pseudo datasets.
As shown in Table \ref{Xling:En}, we find that the translated datasets can bring significantly better performances than the source baseline overall languages,
resulting in an averaged F1 score increase of $51.1-44.4 = 6.7$.
The results demonstrate that corpus translation is one effective way for cross-lingual SRL.
The observation is in line with the previous work for cross-lingual dependency parsing \cite{TiedemannA16,zhang-etal-2019-cross}.
By direct gold-standard corpus translation, the produced pseudo training data can not only remain high-quality SRL annotations
but also capture the language divergences effectively,
which leads to better performance than the source baseline model.

\paragraph{Combining source and pseudo target.}
Further, we combine the pseudo translated target corpus with the source language corpus together to train the target SRL models.
According to the numbers in Table \ref{Xling:En},
we see that further gains can be achieved for all languages,
where the averaged improvement is 55.8-51.1=4.7 (\emph{BASIC} is used for a fair comparison).
Note that since several source sentences are filtered during translation which might be the reason for the gains,
we make a fairer comparison off-the-line by setting $\alpha$=0 (i.e., no sentence filtering).
Similar gains can be achieved still.
Considering that the translated sentences are semantically equal to their counterparts in the gold-standard source,
the possible reasons could be two hands:
(1) the translated sentences may be biased in linguistic expression due to the data-driven translation models,
(2) the discarded conflicted annotations in corpus translation are important, which are complementary to our model.

\begin{table}[!t]
\begin{center}
\resizebox{1.0\columnwidth}{!}{
  \begin{tabular}{c|ccccccc|c}
\hline
Model & ~EN~& ~DE~& ~FR~& ~IT~& ~ES~& ~PT~& ~FI~ & \texttt{Avg}\\
\hline \hline
\multicolumn{9}{c}{\textbf{SRC}}\\ \hline
Emb & 50.3 & 49.2 & 52.4 & 44.9 & 46.7 & 51.0 & 36.4  & 47.3 \\
BERT&	51.8 & 50.6 & 54.0 & 45.3 & 51.3 & 51.8 & 38.1  & 49.0 \\
ELMo&	53.6 & 51.6 & 56.7 & 51.3 & 57.4 & 52.6 & 39.7  & 51.8 \\
\hline\hline
\multicolumn{9}{c}{\textbf{TGT}}\\ \hline
Emb &  56.5 & 51.6 & 55.2 & 47.1 & 50.0 & 53.2 & 40.4  & 50.6 \\
BERT &  59.8 & 55.5 & 57.0 & 52.6 & 54.3 & 56.6 & 44.0  & 54.3 \\
ELMo & 60.7 & 57.8 & 59.9 & 54.8 & 56.7 & 58.8 & 46.9  & 56.5 \\
\hline\hline
\multicolumn{9}{c}{\textbf{SRC \& TGT} (ELMo)} \\ \hline
\emph{BASIC} & 61.9 & 64.8 & 60.3 & 56.4 & 61.1 & 63.1 & 50.7  & 59.8 \\
\emph{PGN} & \textbf{65.7} & \textbf{68.8} & 66.1 & 64.8 & \textbf{68.7} & \textbf{69.2} & \textbf{58.6}  & \textbf{66.0} \\
\hdashline
\emph{MoE} &  63.2 & 67.8 & 63.1 & 62.6 & 65.2 & 67.5 & 54.2  & 63.4 \\
\emph{MAN-MoE} & 64.0 & 68.5 & \textbf{67.2} & \textbf{65.7} & 67.5 & 69.0  & 57.5 & 65.6\\
\hline
\end{tabular}
}
\end{center}
  \caption{Cross-lingual transfer with multiple sources. }
  \label{Mling}
\end{table}

\begin{table}[!t]
\begin{center}
\resizebox{1.0\columnwidth}{!}{
\begin{tabular}{c|ccccccc}
\hline
 Source  & {EN}  & {DE}  & {FR} &  {IT}  & {ES}& {PT} & {FI} \\ \hline \hline
EN &		\multicolumn{1}{|c}{}    &	\multicolumn{1}{c|}{\bf 65.0}& 64.8 & {58.7} & 62.5 & {56.0} & {\bf 54.5}  \\
DE &		\multicolumn{1}{|c}{\bf 63.2}&	 \multicolumn{1}{c|}{}   &	63.9&	60.4&	65.8&	53.4&	50.5 \\ \cline{2-7}
FR &		60.1&	53.7&	\multicolumn{1}{|c}{}    &	63.3&	63.6&	\multicolumn{1}{c|}{62.1}&	51.3 \\
IT &		60.2&	58.9&	\multicolumn{1}{|c}{\bf 65.3}&	    &	65.1&	\multicolumn{1}{c|}{58.6}&	48.6 \\ 
ES &		60.1&	57.3&	\multicolumn{1}{|c}{64.9}&	\bf 64.1&	    & 	\multicolumn{1}{c|}{\bf 67.0}&	50.7 \\  
PT &		57.3&	58.6&	\multicolumn{1}{|c}{65.1}&	63.5&	\bf 67.8&	 \multicolumn{1}{c|}{}    &	40.9 \\ \cline{4-7}
FI &		50.7&	52.1&	64.6&	53.6&	60.3&	51.6 &     \\  \hline
ALL &       65.7&	68.8&	66.1&	64.8&	68.7&	69.2 & 58.6    \\ \hline
\end{tabular}
}
\end{center}
  \caption{The results of bilingual transferring. }
  \label{fine:bilingual}
\end{table}

\paragraph{Language-aware encoder.}
Finally, we investigate the effectiveness of PGN-BiLSTM module, which is exploited to capture language-specific information
when the mixture corpus of both source and target datasets are used for training.
As shown in Table \ref{Xling:En},
we can see that the language-aware encoder by \emph{ PGN} can boost the F1 scores significantly,
achieving an averaged improvement by 60.3-55.8=4.5.
In addition, we report the results of \emph{ MoE} and \emph{ MAN-MoE}, respectively,
which also exploit the language information.
All the results demonstrate the usefulness of language-specific information,
and our \emph{ PGN} model is most effective.

\subsection{Multi-Source Transfer}
Further, we investigate the setting of multi-source transfer learning,
where all other languages except a given target language are used as the source languages,
aiming to study the effectiveness of our translation-based method comprehensively.

\paragraph{Overall performances.}
The results on multiple source SRL transferring are shown in Table \ref{Mling}.
Generally, the results share similar tendencies with the single-source cross-lingual transfer from the source English,
where the multilingual ELMo performs the best,
the SRL models trained on the translated target datasets show better performances than those trained with the source datasets,
and the mixture corpus with both source and target language datasets bring the best performances,
which can be further improved by our final \emph{ PGN} model with language-aware encoders.
We compare the \emph{ PGN} model with the \emph{ MoE} and \emph{ MAN-MoE} as well,
showing slightly better performances, which indicates the effectiveness of the PGN-BiLSTM module.
In addition, we can see that multi-source models outperform the single-source models in all cases,
which is intuitive and consistent with previous studies \cite{lin-etal-2019-choosing}.

\begin{figure}[!t]
\centering
\includegraphics[width=0.65\columnwidth]{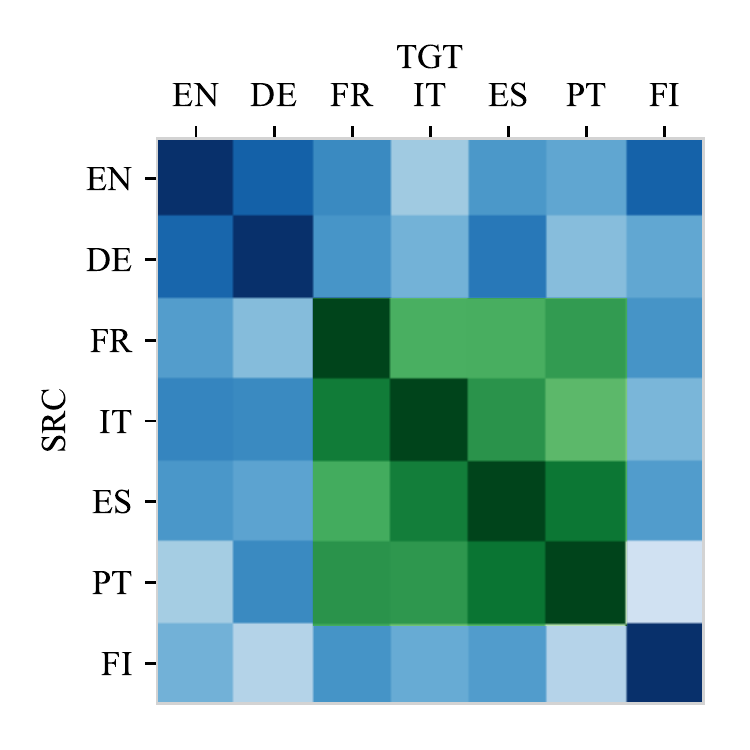}
\caption{
Similarity heatmap of language embeddings for source languages to target languages.
Deeper color indicates higher similarity.
}
\label{similarity}
\end{figure}

\paragraph{Fine-grained bilingual transfer.}
Following, we investigate the individual bilingual SRL transferring by examining the performance of each source-target language pair,
aiming to uncover which language benefits a target most and trying to answer whether all source languages are useful for a target language.
Table \ref{fine:bilingual} shows the results,
where the cross-lingual models are trained on the mixture corpus of the source and translated target datasets.
First, we can see that the languages belonging to a single family can benefit each other greatly,
bringing better performances than the other languages in  the majority of cases (i.e., EN--DE, FR--IT--ES--PT).
Second, the multi-source transfer as indicated by All is able to obtain better performances across all languages,
which further demonstrates its advantages over the single-source transfer.

Further, we look into the \emph{ PGN} model in detail, aiming to understand their capabilities of modeling linguistic-specific information.
We examine it by simply visualizing the language ID embeddings $\bm{e}_{\mathcal{L}}$ of each source-target language pair, respectively,
where their Euclidean distances are depicted.
Intuitively, better performance can be achieved if the distance between the target and the source languages is closer.
Figure \ref{similarity} shows the heatmap matrix.
We can see the overall tendency is highly similar to the results in Table \ref{fine:bilingual},
which is consistent with our intuition.

\subsection{Analysis}

Here we conduct detailed analysis to understand the gains from the translated target datasets.
We select three representative languages for analysis, including German (DE), French (FR) and Finnish (FI),
one language for each family,
and compare four models mainly, including three models (i.e., {\textbf SRC}, {\textbf TGT} and {\textbf SRC \& TGT} with \emph{ PGN})
of the single-source transfer from English and
the final \emph{ PGN} model of multi-source transfer.

\begin{figure}[!t]
\centering
\includegraphics[width=.97\columnwidth]{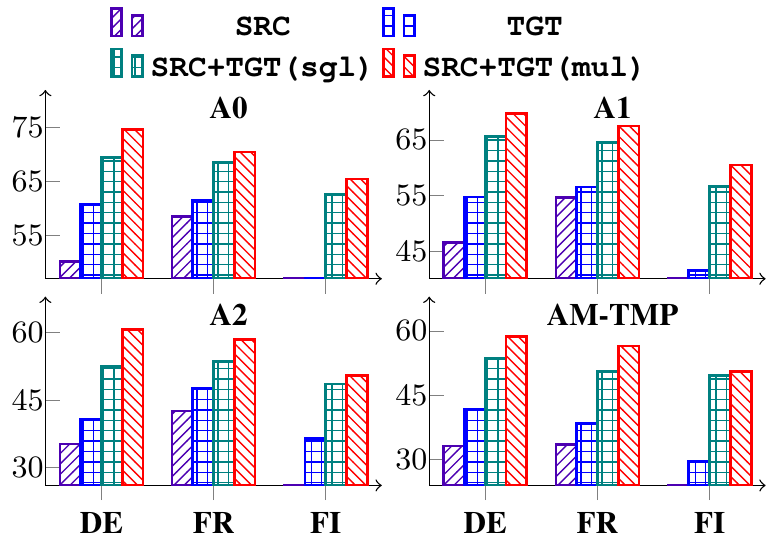}
\caption{
Performances on different argument label.
}
\label{arg label}
\end{figure}

\paragraph{Performances by the SRL roles.}
First, we investigate the cross-lingual SRL performances in terms of SRL Roles.
We select four representative roles for comparison,
including \texttt{A0} (\emph{Agent}), \texttt{A1} (\emph{Patient}), \texttt{A2} (\emph{Instrument, Benefactive, Attribute}) and \texttt{AM-TMP} (\emph{Temporal}),
and report their F1 scores.
Figure \ref{arg label} shows the results.
As a whole, the role \texttt{A0} achieves the best F1 scores across all languages and all models,
\texttt{A1} ranks the second, and \texttt{A2} and \texttt{AM-TMP} are slightly worse.
The tendency could be accounted for by the distribution of these labels,
where \texttt{A0} is the most frequent and \texttt{A2} and \texttt{AM-TMP} have lower frequencies than \texttt{A0} and \texttt{A1}.
The second possible reason could be due to that the majority of the \texttt{A0} and \texttt{A1} words are notional words which could be more easily transferred by cross-lingual models.

\begin{figure}[!t]
\centering
\includegraphics[width=.98\columnwidth]{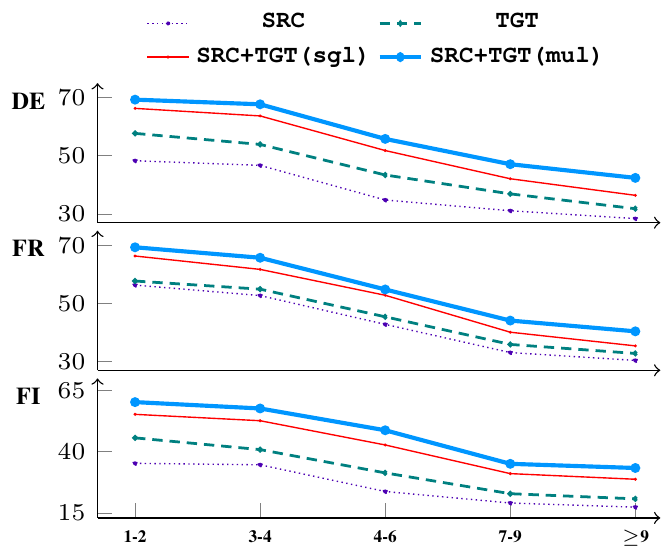}
\caption{
Performances by surface distance between predicates and arguments.
}
\label{prd distance}
\end{figure}

In addition, we can see that the tendencies across different models for all three languages and all labels
are identical, where multi-source transfer performs the best,
single-source {\emph SRC+TGT } ranks the second and our baseline model is the last.
The observation is consistent with the overall tendency,
demonstrating the stability and also further verifying the effectiveness of our proposed models.

\paragraph{Performances by the distances to the predicate.}
Second, we study the SRL performances in terms of the distance to the predicate word.
Intuitively, long-distance relations are more difficult,
thus we expect that the SRL performance would decrease as the distance increases,
as SRL actually detects the relationship between the role words and their predicates.
Figure \ref{prd distance} shows the F1 scores.
First, for all the settings we can see that the SRL performance drops by longer distances,
which confirms our intuition.
In addition, the tendency between different models is the same as the overall results,
demonstrating the effectiveness of our method.

\section{Conclusion}
We proposed a translation-based alternative for cross-lingual SRL.
The key idea is to construct high-quality datasets for the target languages by corpus translation from the gold-standard SRL annotations of the source languages.
In addition, we combined the gold-standard source SRL corpora and the pseudo translated target corpora together to enhance the cross-lingual SRL models.
We investigated cross-lingual SRL models with different kinds of multilingual word representations.
Further, we presented a PGN-BiLSTM encoder to better exploit the mixture corpora of different languages.
Experimental results on the UPB v1.0 dataset show that the translation-based method is an effective method for cross-lingual SRL transferring.
Significant improvements can be achieved by using the translated datasets for all selected languages, including both single-source and multi-source transfer.
Experiment analysis is offered to understand the proposed method in depth.

\section*{Acknowledgments}
This work is supported by the National Natural Science Foundation of China (No.61772378 and 61602160), the National Key Research and Development Program of China (No.2017YFC1200500), 
the Research Foundation of Ministry of Education of China (No.18JZD015), and 
the Major Projects of the National Social Science Foundation of China (No.11\&ZD189).

\bibliography{acl2020}
\bibliographystyle{acl_natbib}

\appendix

\section{Bilingual Transfer by Each Source}

\begin{table*}[!t]
\begin{center}
\resizebox{1.64\columnwidth}{!}{
  \begin{tabular}{c|cccccccc}
\hline
\multirow{2}{*}{  Target} &	\multicolumn{3}{c}{ \texttt{SRC}}&	\multicolumn{3}{c}{\texttt{TGT}}&	\texttt{SRC+TGT}&	\texttt{SRC+TGT} \\  
\cmidrule(r){2-4}\cmidrule(r){5-7}\cmidrule(r){8-8}\cmidrule(r){9-9}
&	Emb	&BERT&	ELMo&	Emb	&BERT&		ELMo&	BASIC+ELMo&	PGN+ELMo \\
\hline
\multicolumn{9}{c}{\quad Source: \textbf{ DE}}\\\hdashline
EN & 47.32 & 51.62 & 52.82 & 55.04 & 59.20 & 60.48 & 61.05 & 63.21\\
FR & 46.00 & 49.94 & 50.99 & 52.37 & 55.77 & 57.02 & 59.91 & 63.90\\
IT & 40.90 & 43.68 & 45.06 & 48.01 & 51.62 & 52.91 & 57.94 & 60.38\\
ES & 39.01 & 42.57 & 43.67 & 49.59 & 52.84 & 53.92 & 60.80 & 65.89\\
PT & 38.25 & 41.73 & 43.07 & 41.44 & 45.76 & 46.94 & 49.14 & 53.40\\
FI & 29.93 & 33.64 & 34.95 & 41.78 & 44.09 & 45.21 & 45.74 & 50.53\\
\hline\hline
\multicolumn{9}{c}{\quad Source: \textbf{ FR}}\\\hdashline
EN & 35.47 & 39.49 & 40.80 & 48.57 & 53.04 & 54.12 & 56.91 & 60.05\\
DE & 40.01 & 43.86 & 45.24 & 41.33 & 45.16 & 46.54 & 50.53 & 53.69\\
IT & 47.12 & 50.06 & 51.33 & 51.45 & 53.38 & 53.62 & 60.31 & 63.34\\
ES & 40.46 & 44.01 & 45.09 & 50.36 & 53.77 & 54.79 & 58.61 & 63.62\\
PT & 44.68 & 47.47 & 48.65 & 52.12 & 55.58 & 56.67 & 59.47 & 62.08\\
FI & 26.44 & 30.92 & 32.07 & 40.71 & 43.97 & 45.05 & 48.76 & 51.31\\
\hline\hline
\multicolumn{9}{c}{\quad Source: \textbf{ IT}}\\\hdashline
EN & 37.07 & 39.49 & 40.96 & 47.10 & 51.26 & 52.40 & 54.05 & 60.13\\
DE & 39.75 & 42.67 & 43.74 & 45.84 & 50.03 & 51.34 & 55.90 & 58.91\\
FR & 47.39 & 50.08 & 51.28 & 54.45 & 57.29 & 58.78 & 60.03 & 65.30\\
ES & 44.29 & 47.92 & 49.14 & 52.09 & 55.68 & 55.08 & 60.56 & 65.09\\
PT & 42.18 & 46.54 & 47.60 & 49.38 & 53.85 & 54.96 & 57.02 & 58.65\\
FI & 31.12 & 33.72 & 35.05 & 40.80 & 43.90 & 44.97 & 46.37 & 48.62\\
\hline\hline
\multicolumn{9}{c}{\quad Source: \textbf{ ES}}\\\hdashline
EN & 41.63 & 44.37 & 45.45 & 48.37 & 52.01 & 53.10 & 55.08 & 60.05\\
DE & 36.32 & 39.65 & 40.73 & 44.21 & 47.90 & 49.37 & 51.11 & 57.27\\
FR & 46.74 & 50.84 & 52.29 & 52.38 & 55.34 & 56.39 & 59.58 & 64.93\\
IT & 41.39 & 45.42 & 46.82 & 50.10 & 53.01 & 54.01 & 58.83 & 64.09\\
PT & 47.52 & 50.46 & 51.68 & 53.44 & 56.49 & 57.54 & 62.30 & 67.01\\
FI & 29.46 & 32.19 & 33.33 & 39.27 & 42.95 & 44.07 & 47.91 & 50.72\\
\hline\hline
\multicolumn{9}{c}{\quad Source: \textbf{ PT}}\\\hdashline
EN & 34.83 & 38.09 & 39.27 & 43.16 & 46.09 & 47.50 & 53.12 & 57.30\\
DE & 37.11 & 41.64 & 42.73 & 46.80 & 49.41 & 50.62 & 55.95 & 58.64\\
FR & 42.05 & 46.28 & 47.61 & 49.07 & 52.78 & 54.15 & 58.64 & 65.12\\
IT & 38.55 & 42.35 & 43.72 & 47.09 & 51.20 & 52.24 & 56.22 & 63.51\\
ES & 39.58 & 44.01 & 45.02 & 46.57 & 50.61 & 52.06 & 60.84 & 67.81\\
FI & 21.54 & 25.01 & 26.24 & 33.53 & 36.91 & 38.03 & 38.50 & 40.90\\
\hline\hline
\multicolumn{9}{c}{\quad Source: \textbf{ FI}}\\\hdashline
EN & 32.99 & 35.99 & 37.38 & 37.83 & 40.48 & 41.65 & 46.80 & 50.70\\
DE & 30.98 & 34.59 & 35.70 & 40.75 & 44.41 & 45.84 & 50.68 & 52.12\\
FR & 39.52 & 43.85 & 44.97 & 48.35 & 50.82 & 52.30 & 56.82 & 64.63\\
IT & 33.82 & 36.39 & 37.66 & 41.81 & 46.01 & 47.07 & 52.61 & 53.65\\
ES & 35.23 & 39.56 & 40.61 & 43.43 & 47.81 & 49.10 & 55.48 & 60.37\\
PT & 27.93 & 31.90 & 33.30 & 33.84 & 38.21 & 39.43 & 47.75 & 51.61\\
\hline\hline
\end{tabular}
}
\end{center}
  \caption{Results of fine-grained bilingual transfer. }
  \label{Xling OTHER detail}
\end{table*}

In the paper, we investigate the individual bilingual SRL transferring of each source target language pair.
We here list the detailed results of the bilingual transfer in Table \ref{Xling OTHER detail}.

\section{Extended SRL Analysis}

\begin{figure}[!t]
\centering
\includegraphics[width=1.0\columnwidth]{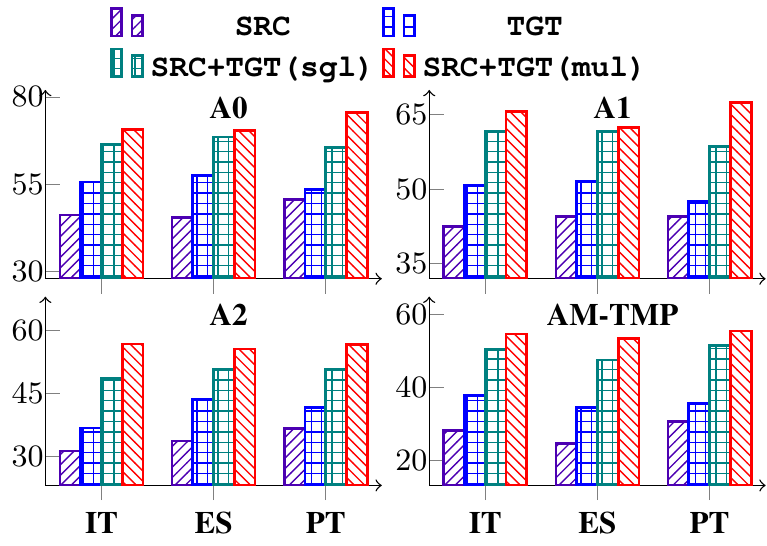}
\caption{
Performances on different argument label.
}
\label{arg label2}
\end{figure}

\begin{figure}[!t]
\centering
\includegraphics[width=1.0\columnwidth]{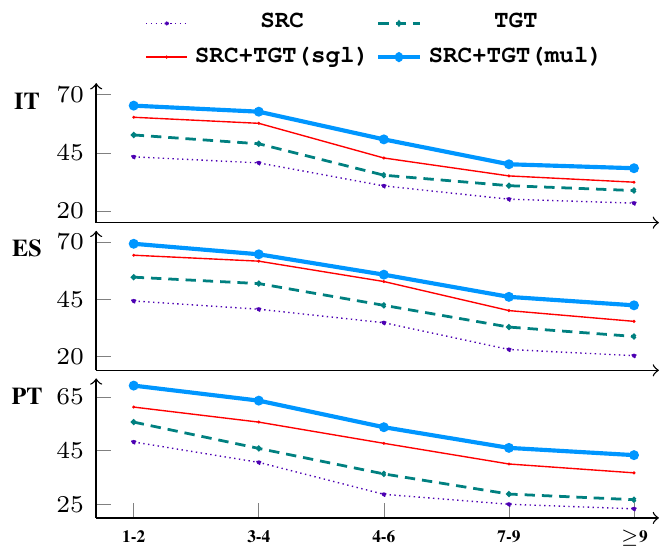}
\caption{
Performances by surface distance between predicates and arguments.
}
\label{prd distance2}
\end{figure}

We conduct detailed analysis on the detailed role labeling and the distance to the predicate, further for the Italian, Spanish and Portuguese languages.
Figures \ref{arg label2} and Figure \ref{prd distance2} show the results.

\end{document}